\documentclass{article}
\usepackage{spconf,amsmath,graphicx}

\title{ST-GRIT: Spatio-Temporal Graph Transformer For Internal Ice Layer Thickness Prediction}
\name{Zesheng Liu$^1$, Maryam Rahnemoonfar$^{1,2,*}$\thanks{This work is supported by NSF BIGDATA awards (IIS-1838230, IIS-2308649), NSF Leadership Class Computing awards (OAC-2139536), IBM, and Amazon.}\thanks{$^*$Correspondence to maryam@lehigh.edu}
\thanks{
© 2025 IEEE. Published in 2025 IEEE International Conference on Image Processing (ICIP), scheduled for 14–17 September 2025 in Anchorage, Alaska, USA. Personal use of this material is permitted. However, permission to reprint/republish this material for advertising or promotional purposes or for creating new collective works for resale or redistribution to servers or lists, or to reuse any copyrighted component of this work in other works, must be obtained from the IEEE. Contact: Manager, Copyrights and Permissions / IEEE Service Center / 445 Hoes Lane / P.O. Box 1331 / Piscataway, NJ 08855-1331, USA. Telephone: + Intl. 908-562-3966.}
\thanks{This version is the accepted manuscript submitted to arXiv. The final version will be published in the Proceedings of ICIP 2025 and available via IEEE Xplore. For citation, please refer to the published version in ICIP 2025.
}}

\address{
$^1$Department of Computer Science and Engineering, Lehigh University, PA, USA\\
$^2$Department of Civil and Environmental Engineering, Lehigh University, PA, USA
}

\begin{document}
%\ninept
%
\maketitle
% \IEEEpeerreviewmaketitle
%
\begin{abstract}
Understanding the thickness and variability of internal ice layers in radar imagery is crucial for monitoring snow accumulation, assessing ice dynamics, and reducing uncertainties in climate models. Radar sensors, capable of penetrating ice, provide detailed radargram images of these internal layers. In this work, we present ST-GRIT, a spatio-temporal graph transformer for ice layer thickness, designed to process these radargrams and capture the spatiotemporal relationships between shallow and deep ice layers. ST-GRIT leverages an inductive geometric graph learning framework to extract local spatial features as feature embeddings and employs a series of temporal and spatial attention blocks separately to model long-range dependencies effectively in both dimensions. Experimental evaluation on radargram data from the Greenland ice sheet demonstrates that ST-GRIT consistently outperforms current state-of-the-art methods and other baseline graph neural networks by achieving lower root mean-squared error. These results highlight the advantages of self-attention mechanisms on graphs over pure graph neural networks, including the ability to handle noise, avoid oversmoothing, and capture long-range dependencies. Moreover, the use of separate spatial and temporal attention blocks allows for distinct and robust learning of spatial relationships and temporal patterns, providing a more comprehensive and effective approach.

% The abstract should appear at the top of the left-hand column of text, about
% 0.5 inch (12 mm) below the title area and no more than 3.125 inches (80 mm) in
% length.  Leave a 0.5 inch (12 mm) space between the end of the abstract and the
% beginning of the main text.  The abstract should contain about 100 to 150
% words, and should be identical to the abstract text submitted electronically
% along with the paper cover sheet.  All manuscripts must be in English, printed
% in black ink.
\end{abstract}
\begin{keywords}
Deep Learning , Remote Sensing , Graph Transformer, Ice Layer, Spatio-Temporal
\end{keywords}
\section{Introduction}

Polar ice sheets are not merely frozen landscapes; they are vital archives of Earth's climate history and key to understanding the pressing challenges of global warming, climate change, and sea level rise. These ice sheets, composed of multiple internal ice layers accumulated over the years, preserve a detailed record of past climate conditions and offer crucial insights into the processes driving today's environmental transformations. Studying these internal ice layers—specifically their thickness and variability—enables researchers to monitor changes in snow accumulation, track sea level rise, and reduce uncertainties in climate models. With rising global temperatures, the loss of polar ice has accelerated dramatically, underscoring the urgency of understanding and mitigating climate impacts. 

The traditional way to study the thickness of internal ice sheets is through onsite ice core, where holes are drilled across the ice sheets and large cylindrical ice samples are manually extracted. However, this approach provides sparse and uneven coverage, limiting the ability to comprehensively understand variations in ice layers across regions. While interpolation of the collected data has seen some success, it introduces uncertainties that can affect the accuracy of climate models. Moreover, this method is confined to easily accessible locations, leaving vast remote areas unexplored, and the drilling process causes physical damage to the ice sheets, raising environmental and conservation concerns.

\begin{figure}[t]
  \centering
  \includegraphics[width=0.25\textwidth]{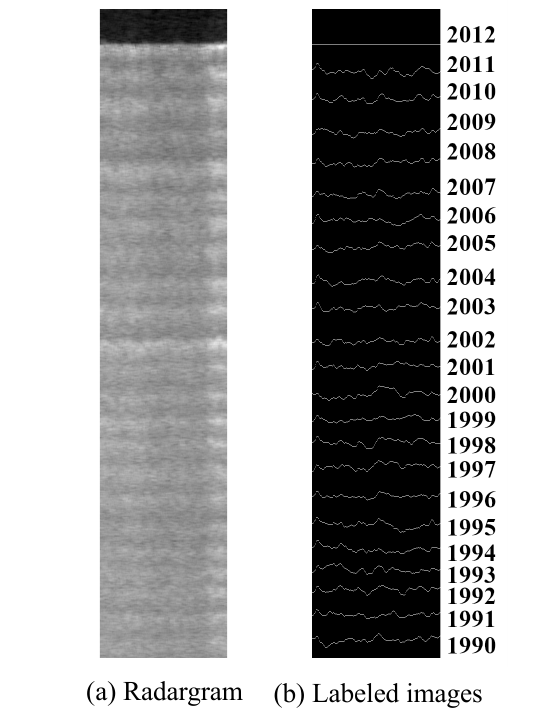}
  \caption{(\textbf{a}) Radargram image (\textbf{b}) Labeled image, where the boundaries of each ice layer is manually labeled out.\label{fig:diagram}}
\end{figure}

In recent years, airborne radar sensors have emerged as a widely used tool for studying ice layers, as they are the main sensors that can traverse thick ice and provide continuous measurements. One example is the snow radar sensor \cite{CReSIS_radar} operated by the Center for Remote Sensing of Ice Sheets (CReSIS)\cite{CReSIS_radar}. By analyzing the strength of the reflected signals \cite{Arnold_2020}, the snow radar sensor can capture the location of internal layers at different depths as various radargrams, such as the one shown in Fig.~\ref{fig:diagram}(\textbf{a}). The radargrams are subsequently labeled, with the annotated lines indicating the boundaries of ice layers formed by the accumulation of snow over successive years, as shown in Fig.~\ref{fig:diagram}(\textbf{b}).

Graph convolutional networks have recently demonstrated remarkable success in various fields, including the prediction of deep ice layer thickness. Zalatan et al. \cite{zalatan_icip}, Liu et al. \cite{liu2024multibranchspatiotemporalgraphneural} introduced approaches that model shallow ice layers as a temporal sequence of spatial graphs, employing diverse spatio-temporal graph neural networks to capture relationships between layers over different years. Although GNNs offer a robust and noise-resilient framework for understanding spatiotemporal patterns in ice layers, their reliance on localized feature aggregation inherently limits their capacity to model more complex temporal dynamics, which often require capturing long-range temporal relationships.

In this paper, we propose ST-GRIT, namely \textbf{S}patio-\textbf{T}emporal \textbf{G}raph t\textbf{R}ansformer for \textbf{I}ce layer \textbf{T}hickness, designed to overcome the limitations of local receptive fields in traditional graph neural networks by incorporating a self-attention mechanism. Compared to traditional GNNs, ST-GRIT incorporates transformer encoders into the geometric deep learning framework, applying it to both spatial and temporal dimensions. The main contributions of this work are:
\begin{itemize}
    \item We developed ST-GRIT, a spatio-temporal graph transformer that learns from the thickness and geographical information of top $p$ internal ice layers and predicts the thickness of underlying $q$ layers.
    
    \item ST-GRIT integrates the GraphSAGE inductive framework to encode spatial node features and incorporates attention mechanisms across both spatial and temporal dimensions. This design enables the model to capture both local and long-range dependencies effectively.
    
    \item We conducted extensive experiments to compare the performance of ST-GRIT with multiple baseline graph neural networks, and the results show that our proposed network consistently has a lower root mean-squared error on predicting the thickness of deep internal ice layers.
\end{itemize}

\section{Related Work}

\subsection{Automatic Internal Ice Layer Boundary Tracking}
Tracking the boundaries of ice layers is a challenging task, as the older deep ice layer may be incomplete or entirely melted. Moreover, the scalability of this task is difficult to handle, as radar sensors typically collect vast amounts of data, demanding efficient processing and analysis algorithms. Various deep learning approaches, such as convolutional neural networks (CNNs) and generative adversarial networks (GANs), have been employed to effectively identify and trace ice layers in radargram images \cite{EisNet,ST_SOLOV2,tiered_ice_segmentation}. Despite achieving some success, these deep learning methods consistently emphasize that noise in the input radargram, combined with the limited availability of high-quality snow radar datasets and annotations, remains a key challenge. Compared with previous convolution-based networks, we represent internal ice layers as independent graphs, and our proposed ST-GRIT focuses on geometric deep learning and attention mechanisms. This approach exhibits greater resilience to noise, ensuring more consistent and reliable performance across inputs of varying radargram quality.

\subsection{Graph Neural Network For Deep Ice Layer Thickness Prediction}

Zalatan et al. \cite{zalatan_icip} proposed to represent each internal ice layer independently and developed a multi-target, adaptive long short-term graph convolutional network (AGCN-LSTM) to estimate the thickness of deep ice layers. By combining a graph convolutional network (GCN) with a long short-term memory (LSTM) structure and incorporating EvolveGCNH \cite{EGCN} as an adaptive layer, their AGCN-LSTM efficiently captures the spatio-temporal dynamics of internal ice layers. Inspired by their idea of using geometric deep learning on internal ice layers, Liu et al. \cite{liu2024multibranchspatiotemporalgraphneural} further improve the accuracy and efficiency of AGCN-LSTM. As the current state-of-the-art method, they applied a multi-branch structure that separates the learning process for spatial and temporal patterns, enabling better weight optimization during the training process. Compared with previous work, our proposed ST-GRIT framework leverages temporal attention blocks to learn the temporal changes across different layers. Unlike traditional graph neural networks, the attention mechanism captures both short-term and long-term patterns, enabling a more precise understanding of temporal dynamics and improving the framework's ability to model complex spatio-temporal relationships.

\begin{figure*}[t]
  \centering
  \includegraphics[width=0.75\textwidth]{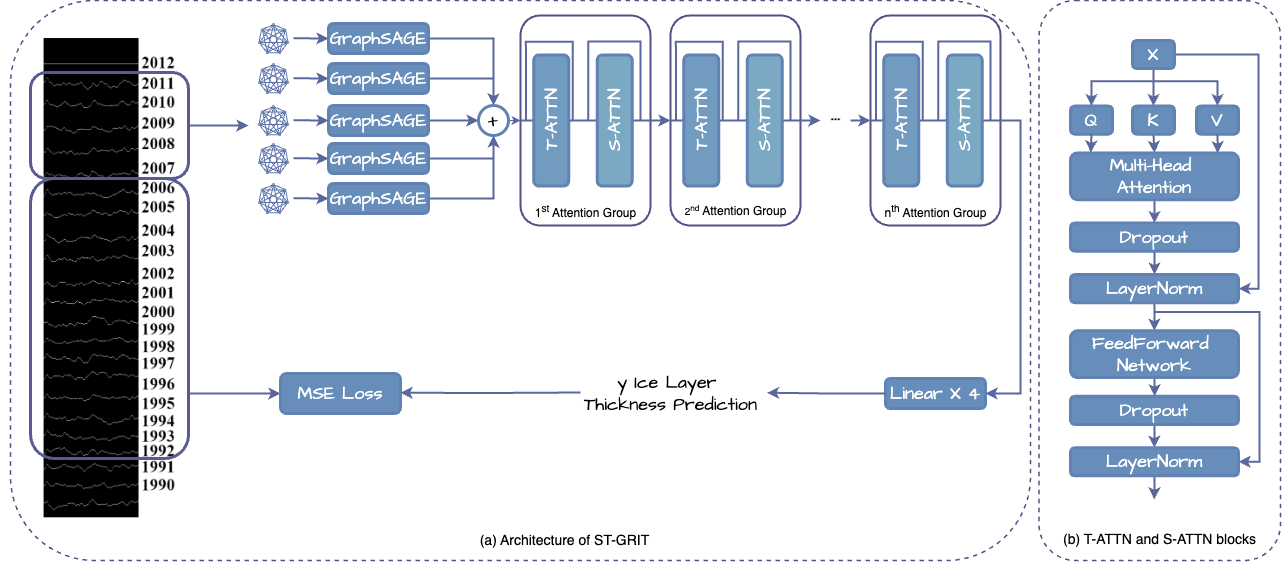}
  \caption{Architecture of our proposed graph transformer network, ST-GRIT.\label{fig:arch}}
\end{figure*}
\section{Radargram Dataset}

We use a radargram dataset captured across the Greenland region 2012. This dataset was collected using an airborne snow radar sensor operated by CReSIS, as part of NASA's Operation IceBridge \cite{Leuschen2011SnowRadar}. Airborne snow radar is one of the most effective platform for assessing the status of internal ice layers, as radar is the primary sensor capable of penetrating thick ice sheets. Internal ice layers are captured as radargram images by measuring the strength of reflected signal \cite{Arnold_2020}, as shown in Fig.~\ref{fig:diagram}(\textbf{a}). Labeled images, like Fig.~\ref{fig:diagram}(\textbf{b}), are generated by manually annotating the boudaries of each layer in the radargram. Using these labeled images, the thickness of each ice layer can be calculated as the difference between the coordinate value of upper and lower boundaries. During radargram acquisition, additional onboard equipment is employed to monitor the aircraft's orientation and motion for simultaneously recording its current latitude and longitude.

\section{Key Designs}

In this work, we developed ST-GRIT, a spatio-temporal graph transformer network that is designed to learn the relationships between the shallow and deeper ice layers, aiming to learn from the geographical and thickness information of the top $p$ ice layers and estimate the thickness of $q$ layers beneath. The key innovation of ST-GRIT lies in its use of self-attention mechanisms, enabling more robust learning in both spatial and temporal dimensions. This approach addresses several key limitations of CNNs, RNNs, and traditional GNNs, such as restricted local receptive fields, vanishing gradients, and oversmoothing. By dynamically capturing long-range dependencies, self-attention allows ST-GRIT to handle noise effectively while preserving critical spatial and temporal patterns. Furthermore, using separate attention encoders for spatial and temporal dimensions ensures that the unique characteristics of spatial relationships (e.g., geographical adjacency) and temporal dependencies (e.g., annual variations) are learned independently. This separation prevents the conflation of features, which could otherwise degrade performance.

ST-GRIT takes temporal sequences of $p$ spatial graphs as the network inputs. As shown in Fig.~\ref{fig:arch}(\textbf{a}), the network starts with a GNN part composed of five GraphSAGE blocks, aiming to learn the spatial patterns within each graph and concatenate them as feature embeddings for the temporal attention blocks. The learned embeddings are then passed into $n$ groups of temporal and spatial attention blocks. Skip connections are used to connect the input and output of each attention block, which ensures a more robust and stable learning process. In the end, four linear layers are used as a decoder to make the final prediction on the $q$ layers' thickness. 

\subsection{GraphSAGE Inductive Framework}
In ST-GRIT, we employed five independent GraphSAGE models to capture the spatial patterns within each internal layer and generate feature embeddings. Unlike Graph Convolution Network (GCN) \cite{kipf2017_GCN}, GraphSAGE \cite{hamilton2018inductive_graphsage} is an inductive framework specifically designed to produce node embeddings for previously unseen data by utilizing local neighbor sampling and feature aggregation techniques \cite{ZHOU202057_Review}. For an unseen node $i$ with node feature matrix $x_i$, the learned node embedding calculated by GraphSAGE is defined as:
\begin{equation}
\textbf{x}'_i = \textbf{W}_1 \textbf{x}_i + \textbf{W}_2 \cdot \textit{mean}_{j \in \mathcal{N}(i)} \textbf{x}_j
\end{equation}
where $\textbf{W}_1$ and $\textbf{W}_2$ are learnable weights matrix, $\mathcal{N}(i)$ is the neighbor list of node $i$ and usually includes neighbors with different depth, $\textbf{x}_j$ is the node feature matrix for neighbor node, and $\textit{mean}$ is the aggregation function. GraphSAGE offers key advantages over GCN for modeling ice layer graphs. Unlike GCN, which depends on a fixed global graph Laplacian and is less adaptable to varying graph structures, GraphSAGE uses a spatial aggregation approach that supports arbitrary and evolving topologies. This makes it well-suited for our setting, where graph structures change across years and regions. Importantly, GraphSAGE applies separate transformations to root and neighbor features using distinct weight matrices, allowing nodes to retain their individual geophysical identity while incorporating contextual information. This design not only enhances generalization across diverse graphs but also helps preserve critical node-specific features—such as local thickness—early in the network, reducing the risk of over-smoothing before attention-based processing.

\subsection{Spatial And Temporal Attention Layer}
In the ST-GRIT encoder, we incorporated a series of attention groups composed of spatial and temporal attention blocks based on the standard multi-head attention mechanism introduced by Vaswani et al. \cite{vaswani2017attention}. Each attention block stacks multiple scaled dot-product attention, enabling the model to effectively capture temporal and spatial dependencies and enhance overall performance. For input feature embedding $X$, queries, keys, and values for each head are generated as:
\begin{equation}
    Q_i=XW_i^Q, K_i=XW_i^K, V_i=XW_i^V
\end{equation}
where $W_i^Q, W_i^K, W_i^V$ are learnable weights. Attention calculations are performed independently in each head with its own query, key, and value matrix: 
\begin{equation}
Attention(Q_i,K_i,V_i)=softmax(\frac{Q_iK_i^T}{\sqrt{d_k}})V_i
\end{equation}
where $d_k$ is the dimension of the key vector in each head. All these independent results are finally concatenated together to produce the multi-head attention output:
\begin{equation}
    MultiHead(X) = Concat(Head_1, Head_2,...,Head_n)W^o
\end{equation}
where $Head_i = Attention(Q_i, K_i, V_i)$ and $W^o$ is a learnable matrix used to concatenate the result of each head. As shown in Fig.~\ref{fig:arch}(\textbf{b}), besides the multi-head attention layer, our temporal attention also contains dropout, layer normalization, feedforward network, and skip connections, which are common components of transformer encoder. For ST-GRIT, we use 8 heads in total. Inspired by Liu et al. \cite{liu2024multibranchspatiotemporalgraphneural}, which showed that separating the learning of spatial and temporal patterns improves overall accuracy, we follow a similar approach in ST-GRIT by introducing distinct spatial and temporal attention blocks. The primary difference between these blocks lies in the dimension along which the self-attention score is computed. During implementation, necessary transpose operations are applied to ensure that self-attention is computed along the intended dimension.

\section{Experiments and Results}
\begin{figure*}[!t]
  \centering
  \includegraphics[width=0.73\textwidth]{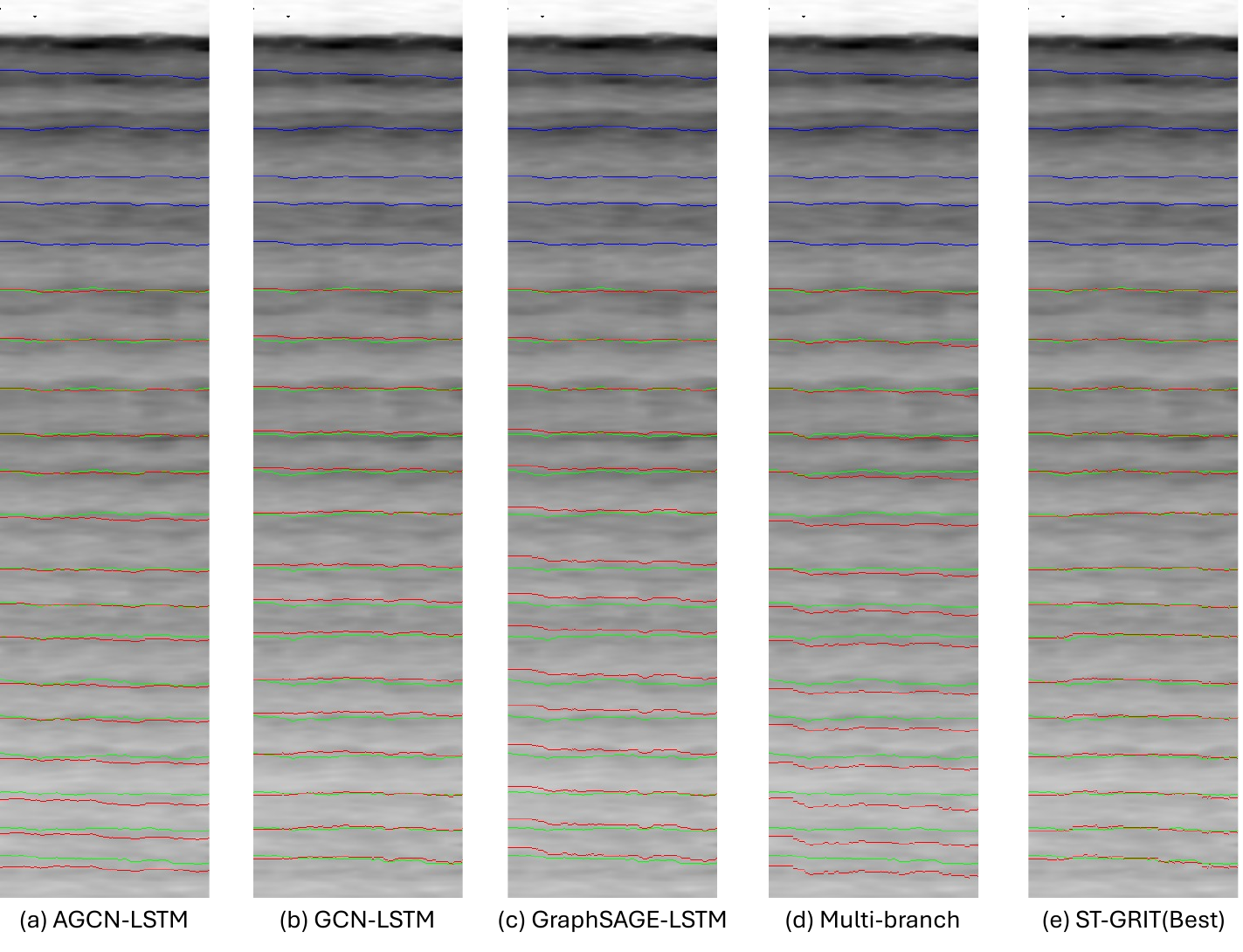}
  \caption{Comparison between ST-GRIT predictions and other baseline models. The blue line is used to generate the graphs. The green line is the groundtruth (manually-labeled ice layers) and the red line is the ST-GRIT prediction.\label{fig:qualitative}}
\end{figure*}

\begin{table}[!t]
    \caption{Experiment results of GCN-LSTM, GraphSAGE-LSTM, Multi-branch GNN, and proposed ST-GRIT.}
    \begin{center}
    \scalebox{0.65}{
    \begin{tabular}{ccc}
    \hline
\textbf{Model}                 & \textbf{RMSE}  &  \textbf{Computation Time (Seconds)}\\ 
\hline
AGCN-LSTM \cite{zalatan_icip}             & 3.4808 $\pm$ 0.0397 & 9404 \\
GCN-LSTM \cite{zalatan_icip}              & 3.1745 $\pm$ 0.1045 & 7441  \\
GraphSAGE-LSTM \cite{liu2024multibranchspatiotemporalgraphneural} & 3.3837 $\pm$ 0.1102 & 4579 \\
Multi-branch \cite{liu2024multibranchspatiotemporalgraphneural} & 3.1087 $\pm$ 0.0555 & 987\\
ST-GRIT(1 Group) & \textbf{2.9727 $\pm$ 0.0475} & 1989\\
ST-GRIT(2 Groups, Best Case) & \textbf{2.8866 $\pm$ 0.0569} & 2459 \\
\hline
\end{tabular}
}
\end{center}
    \label{table:OverallResults}
\end{table}

\subsection{Data Preprocessing and Graph Generation}
We test the proposed ST-GRIT in a specific case with $p=5$ and $q=15$, aiming to use the geographical and thickness information of the top 5 ice layers (formed in 2007-2011) to predict the thickness for the underlying 15 layers (formed in 1992-2006). Given that the number of ice layers can vary by location and some layers may be incomplete, we perform data pre-processing to ensure high data quality by excluding radargram images with fewer than 20 complete ice layers. This results in a total of 1,660 images, which are then divided into training, validation, and test sets in a $3:1:1$ ratio.

The graph dataset is generated by converting each ground-truth radargram into a temporal sequence of five spatial graphs, with each spatial graph representing a single ice layer independently. Each spatial graph is composed of 256 nodes, all of which are fully connected by undirected edges. Edge weights are computed as: 
\begin{equation}
    w_{i,j}  = \frac{1}{2\arcsin{(hav(\phi_j-\phi_i)+\cos{\phi_i}\cos{\phi_j}hav(\lambda_j-\lambda_i))}}
\end{equation}
where $i, j$ can be any node, $w_{i, j}$ is the computed edge weights, $\phi, \lambda$ are the latitude and longitude of nodes, and $hav(\theta) = \sin^2{(\frac{\theta}{2})}$. Each node will have three node features: latitude, longitude and thickness.

\subsection{Training Details}
We compare our proposed ST-GRIT with the current state-of-the-art multi-branch spatio-temporal graph neural network \cite{liu2024multibranchspatiotemporalgraphneural}, together with several baseline graph neural networks, including AGCN-LSTM \cite{zalatan_icip}, GCN-LSTM \cite{zalatan_icip}, and SAGE-LSTM \cite{liu2024multibranchspatiotemporalgraphneural}. All the networks are trained on the same machine with 8 NVIDIA A5000 GPUs. Mean squared error (MSE) loss is used as the training loss function for all the networks. For all the baselines, we used 0.01 as the initial learning rate and used a step learning rate scheduler that halves the learning rate every 75 epochs. For ST-GRIT with a different number of attention groups, we used 0.0005 as an initial learning rate and used an adaptive learning rate scheduler that halved the rate after 24 epochs of no validation loss improvement. All the networks are trained for 450 epochs to ensure full convergence. To minimize potential bias, we generated five distinct versions of the training, validation, and testing datasets by applying different random permutations to the entire set of 1,660 valid images prior to splitting. Each graph neural network was subsequently trained on all five dataset versions.

\subsection{Results and Analysis}
For each dataset version (training, validation, testing), we computed the RMSE between predicted and ground truth layer thickness on the testing dataset for deeper layers. The mean and standard deviation of RMSE across five versions are summarized in Table \ref{table:OverallResults}. Compared to pure graph neural networks, integrating attention mechanisms and transformer blocks greatly improves the model's ability to capture both spatial and temporal long-range dependencies. Our experiments show that $n=2$, i.e., utilizing two attention groups, achieves the best performance for ST-GRIT. Fig.~\ref{fig:qualitative} shows the comparison between ST-GRIT prediction and other baseline models. We observed that the lower RMSE of ST-GRIT is attributed to its improved ability to effectively reduce error accumulation towards those most deeper layers.

In terms of computational time, although our proposed ST-GRIT is slower than the multi-branch GNN, it is important to note that the multi-branch GNN was specifically optimized for maximum efficiency, whereas ST-GRIT is designed with a focus on achieving higher predictive accuracy. Despite the additional computational cost, the training and inference times of ST-GRIT remain within an acceptable range for our application. More importantly, in the context of ice layer thickness prediction, accuracy takes precedence over speed, as reducing uncertainty in the predictions is critical for improving the reliability of downstream scientific analyses and decision-making. Therefore, we view the trade-off between accuracy and efficiency as a worthwhile compromise that aligns with the overall goals of our work.
\section{Conclusion}

In this work, we introduced ST-GRIT, a spatio-temporal graph transformer specifically designed to predict the thickness of deeper internal ice layers in ice sheets. ST-GRIT leverages the GraphSAGE inductive framework to extract spatial feature embeddings and employs multiple temporal and spatial attention blocks to capture long-range dependencies in both dimensions, addressing limitations of traditional GNNs. We evaluated ST-GRIT on a specific use case, utilizing data from ice layers of the Greenland ice sheet formed between 2007 and 2011 to predict the thickness of layers formed from 1992 to 2006. Notably, ST-GRIT demonstrates great flexibility by accommodating a varying number of ice layers and radargrams of different sizes. Experiment results show that ST-GRIT consistently outperforms baseline GNNs with a lower RMSE error, highlighting how integrating an attention mechanism into geometric learning frameworks enables more effective and robust modeling of spatiotemporal dependencies. In future work, we plan to optimize ST-GRIT for improved computational efficiency and faster inference. Additionally, we will develop more advanced data preprocessing techniques to incorporate radargrams with incomplete or noisy ice layer information, enabling broader applicability across diverse datasets.

% \vfill\pagebreak

% -------------------------------------------------------------------------
\bibliographystyle{IEEEbib}
\bibliography{refs}

\end{document}